\documentclass{article}

\usepackage{microtype}
\usepackage{graphicx}
\usepackage{subcaption}
\usepackage{booktabs} % for professional tables
\usepackage{placeins}

\usepackage{hyperref}

\usepackage[preprint]{icml2026}

\usepackage{amsmath}
\usepackage{amssymb}
\usepackage{mathtools}
\usepackage{amsthm}

\usepackage[capitalize,noabbrev]{cleveref}

\theoremstyle{plain}

\theoremstyle{definition}

\theoremstyle{remark}

\usepackage[textsize=tiny]{todonotes}

\usepackage{soul}
\usepackage{comment}

\icmltitlerunning{Reasoning by Commented Code for Table Question Answering}

\begin{document}

\twocolumn[
  \icmltitle{Reasoning by Commented Code for Table Question Answering}



  \icmlsetsymbol{equal}{*}

  \begin{icmlauthorlist}
    \icmlauthor{Seho Pyo}{snu-ds}
    \icmlauthor{Jiheon Seok}{snu-ds}
    \icmlauthor{Jaejin Lee}{snu-ds,snu-cs}
  \end{icmlauthorlist}
  \icmlaffiliation{snu-ds}{
    Department of Data Science, Seoul National University, Seoul, Republic of Korea
  }
  \icmlaffiliation{snu-cs}{
    Department of Computer Science and Engineering, Seoul National University, Seoul, Republic of Korea
  }

  \icmlcorrespondingauthor{Seho Pyo}{hori14@snu.ac.kr}
  \icmlcorrespondingauthor{Jiheon Seok}{jh980120@snu.ac.kr}
  \icmlcorrespondingauthor{Jaejin Lee}{jaejin@snu.ac.kr}

  \icmlkeywords{Machine Learning, ICML}

  \vskip 0.3in
]



\printAffiliationsAndNotice{}  % no special notice (required even if empty)

\begin{abstract}
Table Question Answering (TableQA) poses a significant challenge for large language models (LLMs) because conventional linearization of tables often disrupts the two-dimensional relationships intrinsic to structured data. Existing methods, which depend on end-to-end answer generation or single-line program queries, typically exhibit limited numerical accuracy and reduced interpretability. This work introduces a commented, step-by-step code-generation framework that incorporates explicit reasoning into the Python program-generation process. The approach decomposes TableQA reasoning into multi-line executable programs with concise natural language comments, thereby promoting clearer reasoning and increasing the likelihood of generating correct code. On the WikiTableQuestions benchmark, the proposed method achieves 70.9\% accuracy using Qwen2.5-Coder-7B-Instruct, surpassing the Repanda baseline (67.6\%). Integrating the proposed framework with a robust end-to-end TableQA model via a lightweight answer-selection mechanism yields further improvements. This combined approach achieves up to 84.3\% accuracy on the WikiTableQuestions benchmark.

\end{abstract}

\section{Introduction}

Large language models (LLMs) have demonstrated significant capabilities in natural language understanding and reasoning across diverse tasks, including question answering, logical inference, and code generation~\citep{brown2020language,wei2022chainofthought,openai2023gpt4,gao2023pal,liu2024potplus,zhou2025planandsolve}. Nevertheless, the core mechanism of LLMs, namely autoregressive next-token prediction, presents fundamental challenges for reasoning over structured data such as tables~\citep{xu2023progprompt,guo2023neural,zhang2024hierarch}. Because tables encode information in a two-dimensional format that does not correspond to the linear token processing of LLMs, effective table reasoning remains difficult~\citep{herzig2020tapas, chen2020open, liu2023rethinking}.

% Unlike unstructured text, tables encode information in an explicit two-dimensional layout, where rows and columns jointly define semantic relations.
% This structural nature is not naturally aligned with the linear token sequences processed by LLMs, making table-based reasoning particularly difficult.

Despite these challenges, the capacity of large language models (LLMs) to reason over tabular data is of considerable practical significance. Tabular data is prevalent in real-world contexts, including financial reports, scientific datasets, business analytics, and operational logs~\citep{pasupat2015wtq,herzig2020tapas}. As a result, Table Question Answering (TableQA) research has investigated text-to-SQL, end-to-end methods, and program-based methods, each of which exhibits limitations when applied to complex, noisy tables~\citep{liu2024compositional,patel2024executionguided,patnaik2024cabinet}.

% Consequently, a large body of work has explored table question answering (TableQA), including text-to-SQL translation, end-to-end table reasoning, and program-based approaches that generate executable code.
% While these methods have achieved promising results, each comes with notable limitations when applied to realistic, noisy tables.

Text-to-SQL models convert natural language questions into SQL queries, facilitating precise numerical computation~\citep{zhong2017seq2sql,yu2018spider}. These models generally assume homogeneous datatypes within columns, an assumption frequently violated in real-world tables where numeric values are often mixed with symbols, annotations, or inconsistent formatting. End-to-end TableQA models generate answers through autoregressive reasoning; however, they face scalability challenges as table size increases and are susceptible to context-related issues such as the lost-in-the-middle phenomenon, where performance declines when relevant information appears in the middle of a lengthy context~\citep{liu2023lost,kim2025longtable}. Additionally, these models cannot guarantee numerical exactness because arithmetic operations are implicitly simulated through token generation. Program-based approaches using Python or Pandas provide greater numerical reliability, but previous work often compresses computations into a single expression or omits explicit reasoning steps, thereby reducing interpretability and hindering error analysis~\citep{liu2024compositional,liu2024potplus}.

\begin{figure*}[!t]
    \centering
    \includegraphics[width=0.8\linewidth]{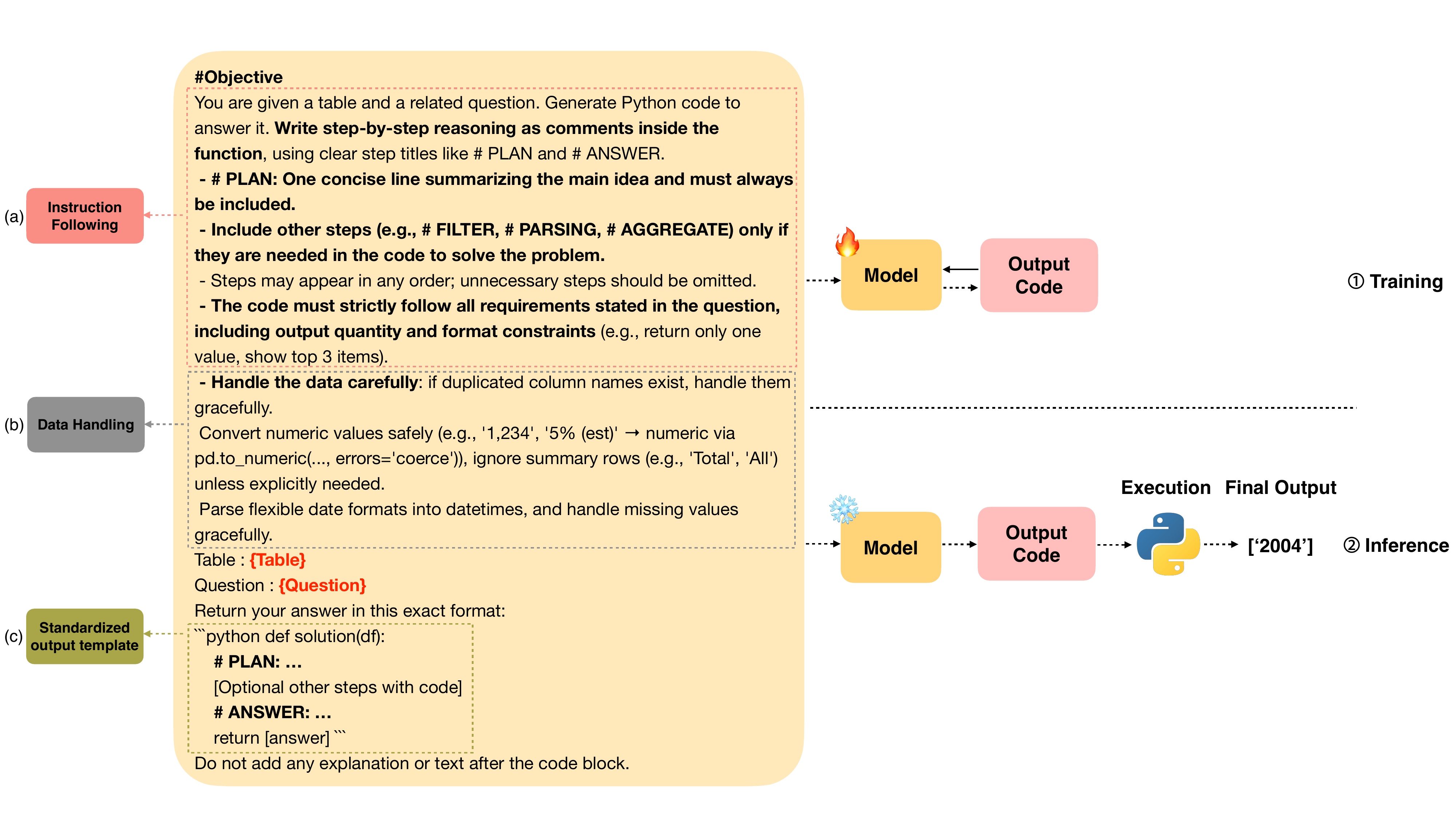}
    \caption{Overview of the proposed framework.}
    \label{fig:framework}
\end{figure*}

This paper introduces a code-based, commented reasoning framework for TableQA that grounds the reasoning process directly in executable Pandas code. Rather than relying on free-form natural-language explanations or implicit program traces, the proposed approach requires the model to generate a multi-line Python function in which each major Pandas dataframe operation is explicitly preceded by a structured reasoning comment~\citep{chen2024toolaug,wang2024codepolicy}. A required high-level planning step, indicated by "\# PLAN," outlines the overall strategy, and subsequent comments align step-wise reasoning with specific Pandas operations. This design maintains the numerical reliability and scalability of code execution while enhancing the transparency and interpretability of the reasoning process.

Figure~\ref{fig:framework} demonstrates that both training and inference use a unified program-generation pipeline~\citep{zhou2025apiguided,park2024contextexec}. When provided with a table and a natural-language question, the model generates a single Python function that integrates high-level planning, stepwise reasoning, and executable Pandas operations. During inference, this program is executed on the relevant Pandas dataframe to yield the final answer. The same execution-based process is applied during training to construct and validate supervision data based on the comment generated by the LLM (e.g., \# PLAN, \# PARSING, amd \# AGGREGATE).

The proposed approach is evaluated on the WikiTableQuestions benchmark~\citep{pasupat2015wtq}, with particular attention to scenarios that demand reliable numerical computation and robust processing of noisy tabular data.
The framework undergoes execution-based evaluation and can be integrated with advanced end-to-end TableQA models through an answer selection module, thereby leveraging complementary strengths without altering the underlying models~\citep{lee2024modular,lin2024multitool,liu2024tabexec}.

The contributions of this paper are summarized as follows:
\begin{itemize}
    \item We introduce a commented reasoning framework that aligns step-wise table reasoning with executable Pandas code, thereby enabling interpretable and reliable TableQA.
    \item We design a structured instruction format to enforce explicit planning, instruction following, and robust data handling within a single-pass program-generation setting.
    \item We demonstrate that grounding reasoning in executable code improves numerical reliability and scalability on realistic tables, and can be effectively combined with end-to-end TableQA models.
\end{itemize}

\section{Related Work}

\subsection{Text-to-SQL}
Text-to-SQL approaches translate natural language questions into executable SQL queries~\citep{zhong2017seq2sql,yu2018spider}. Although these methods are capable of performing numerical operations, they typically assume that each column contains homogeneous data types. In practice, real-world datasets such as WikiTQ~\citep{pasupat2015wtq} often include tables with noisy or mixed data, including numeric values surrounded by parentheses, annotations, or symbols. Under these circumstances, conventional text-to-SQL systems often treat entire columns as strings, hindering reliable numerical computation and limiting their effectiveness when applied to heterogeneous or unstructured tables~\citep{gupta2024sqlinf,wang2024tapexplus}.

Recent approaches, such as Tabsqlify~\citep{zhu2024tabsqlify}, address this issue by first generating SQL queries to construct a relevant sub-table, followed by applying a downstream reasoning model to answer questions over the reduced table. Nevertheless, the intermediate SQL generation step continues to rely on the core assumptions of text-to-SQL, including the necessity for consistent column datatypes to accurately filter, aggregate, or compare values. Consequently, the sub-table construction process remains vulnerable when applied to noisy real-world tables.

In contrast, the proposed framework directly generates Pandas code with explicit reasoning comments, enabling robust numerical operations even when mixed or noisy data types are present. By grounding reasoning in executable code rather than assuming clean tabular schemas, this method can naturally address the variability and irregularities commonly observed in real-world table data.

\subsection{Code-based Table Question Answering}
Previous approaches~\cite{wang2024chainoftable, zhang2025repanda, wang2024accurate} have investigated table reasoning using executable codes. For instance, Chain of Table~\citep{wang2024chainoftable} frames reasoning as a sequence of discrete table operations. At each step, the model selects an operation and its arguments, applies the operation to generate an intermediate sub-table, and continues until an end condition is met. The resulting sub-table and the question are subsequently provided to a language model to generate the answer. Although this method offers an explicit, interpretable reasoning process, it requires multiple rounds of inference and table transformations, resulting in significant token and computational overhead ~\citep{liu2024compositional}.

Repanda~\citep{zhang2025repanda} offers a more token-efficient approach by directly generating a Pandas query from a given table and question. This method is grounded in the principle of explicit reasoning via executable programs, with the generated Pandas expression serving as an implicit reasoning trace. By condensing the reasoning process into a single Pandas expression, Repanda demonstrates high efficiency in both generation and execution. However, due to its output design, Repanda restricts the model to generating a single-line pandas query without comments, explanations, or multi-line formatting. Consequently, although the approach seeks to incorporate structural reasoning, the output reveals only the final dataframe operation rather than a stepwise decomposition of the reasoning process~\citep{liu2024potplus}. Intermediate decisions, such as filtering, parsing, or aggregation, are not explicitly represented, limiting interpretability and hindering comprehensive error analysis~\citep{singh2025tablebias}.

Planner-based approaches, such as TabLaP~\citep{wang2024accurate}, use large language models to initially generate natural-language reasoning or answers, followed by a separate Python program that reproduces or verifies the predicted result. Although this design incorporates executable components, the generated code is only loosely integrated with the reasoning process and primarily serves as a post hoc verification mechanism rather than an explicit reasoning scaffold.

The proposed approach offers an intermediate solution between existing Python-based paradigms that prioritize either iterative, multi-step execution or compact, yet implicit, program generation. It also addresses the weak integration between reasoning and execution observed in planner-based frameworks. Similar to Repanda, this method produces a single executable program in one inference pass, thereby maintaining high token efficiency. However, unlike both Repanda and planner-style methods, it explicitly decomposes table reasoning into multiple, sequential dataframe operations within a multi-line Python function. This design makes reasoning an explicit part of the Python code-generation process. Each primary operation is introduced by a corresponding reasoning comment (e.g., \texttt{\# PLAN}, \texttt{\# FILTER}, \texttt{\# AGGREGATE}), thereby making the reasoning process explicit and closely aligned with execution. By enforcing step-wise decomposition without iterative inference, our framework retains the efficiency of single-pass program generation while offering the transparency and interpretability characteristic of multi-step reasoning approaches. This structure facilitates reliable debugging and inspection of intermediate decisions in complex table reasoning scenarios.

\subsection{End-to-end Table Question Answering}
End-to-end TableQA methods generate answers from a table and a question using autoregressive generation, eliminating the need for explicit program execution. Recent reasoning-centric models, such as Table-R1~\citep{wang2025tabler1}, demonstrate strong performance on benchmarks including WikiTQ. However, these models retain the fundamental limitations of large language models. As table size increases, end-to-end TableQA models become susceptible to context-related issues, such as the lost-in-the-middle phenomenon, which complicates the retrieval of reliable evidence from large tables~\citep{liu2023lost,kim2025longtable}. Furthermore, these models perform numerical operations via token generation rather than explicit computation, which fails to ensure accuracy in aggregation or arithmetic reasoning~\citep{nan2021fetaqa}.

In contrast, the proposed framework grounds reasoning in executable pandas code, ensuring that numerical operations are performed deterministically on structured Pandas dataframes. Delegating computation to a symbolic execution engine mitigates the scalability and numerical reliability challenges associated with purely generative reasoning~\citep{chen2024toolaug,wang2024codepolicy}. Additionally, explicit step-wise reasoning is maintained through commented code, which supports both interpretability and robust execution on larger or noisier tables.

\section{Methods}

% \begin{figure}[t]
%     \centering
%     \includegraphics[width=\columnwidth]{Figure/input_output_format.png}
%     \caption{Unified input–output format used for both training and inference}
%     \label{fig:input_output}
% \end{figure}

\begin{figure*}[!t]
\centering
\begin{minipage}{0.8\linewidth}
    \centering
    \includegraphics[
        width=\linewidth,
        trim=0cm 10.0cm 0cm 8.0cm,
        clip
    ]{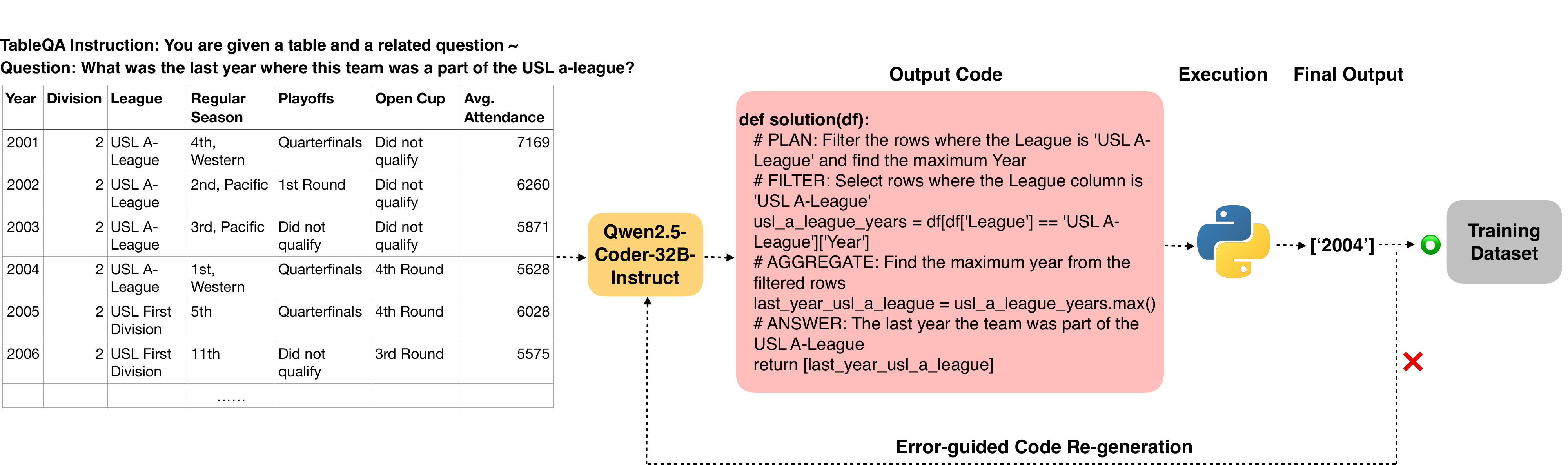}
    \caption{Overview of dataset generation.}
    \label{fig:dataset_generation}
%\end{figure*}
\end{minipage}\\

\vspace{\baselineskip}
\begin{minipage}{0.8\linewidth}
%\begin{figure*}[!t]
    \centering
    \includegraphics[width=\textwidth]{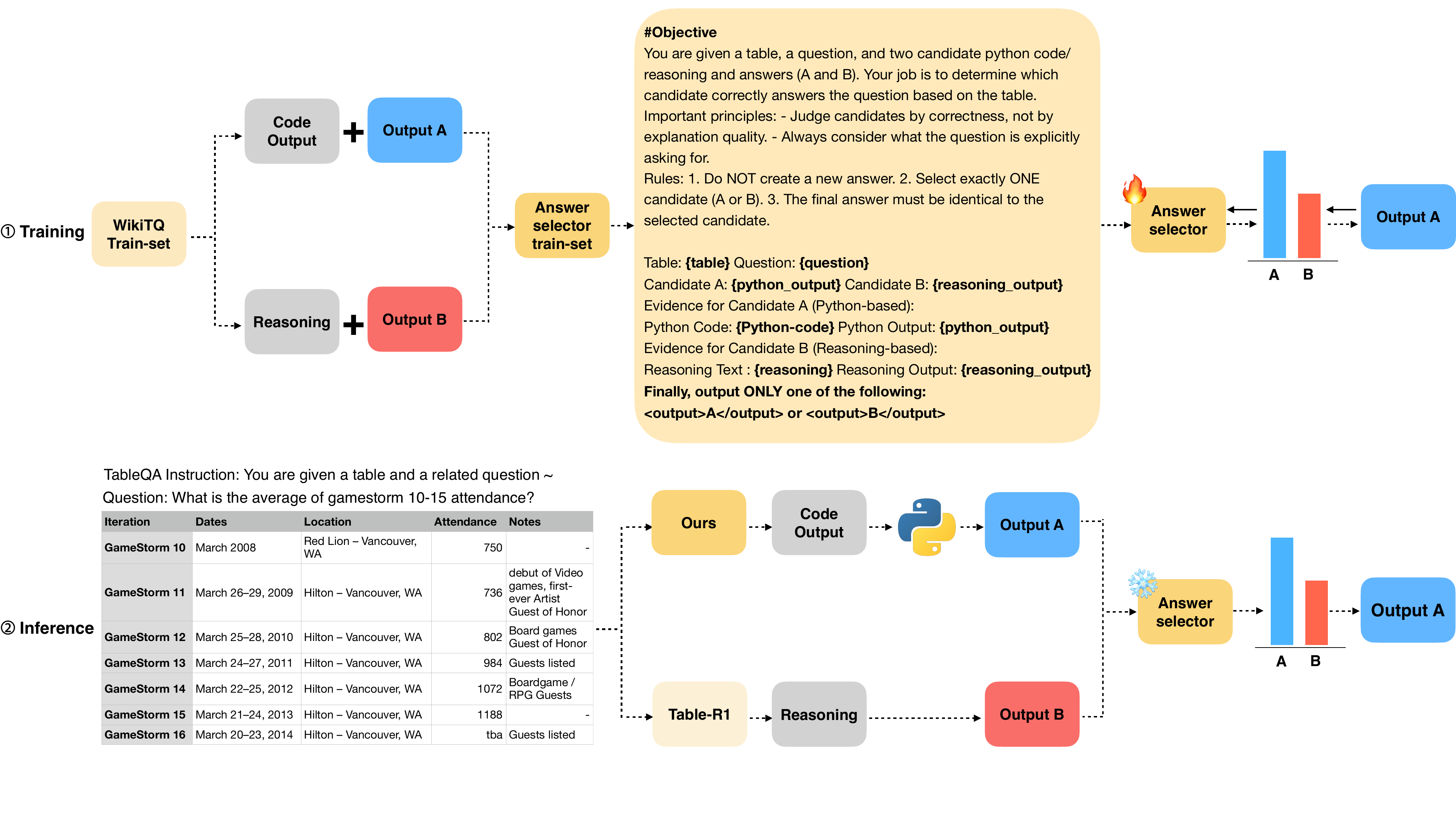}
    \caption{Overview of the answer selector}
    \label{fig:Answer selector}
\end{minipage}

\end{figure*}

%\subsection{Overview}
This paper introduces Table QA as a step-by-step process enabling a large language model (LLM) to generate programs. When provided with an instruction, a table, and a question, the model applies intermediate reasoning to produce an executable Python function. The method decomposes table operations into a sequence of explicit Pandas dataframe steps, each annotated with a concise reasoning comment, as illustrated in Figure~\ref{fig:framework}.

Previous approaches to table question answering have typically treated end-to-end methods and program-based methods as distinct paradigms. End-to-end models generate expressive reasoning traces that are not directly executable, while program-based methods often yield executable code with reasoning that is implicit or external to the program.

Effective table reasoning requires closer integration between reasoning and execution, particularly during program generation. Instead of appending reasoning to code as a post hoc explanation or using code solely for verification, this approach restructures code generation to explicitly decompose programs into step-by-step operations, each paired with a concise reasoning comment.

This design embeds reasoning directly into the structure of the executable program, aligning natural-language explanations with symbolic dataframe operations. Consequently, reasoning becomes an integral component of the program representation, supporting both robust execution and interpretable intermediate structure.

Formally, given a table $T$ and a question $q$, the model generates a commented
program
\[
p = \{(c_1, o_1), \dots, (c_K, o_K)\},
\]
where each pair $(c_i, o_i)$ consists of a natural language reasoning comment $c_i$
and an executable Pandas dataframe operation $o_i$. The program is executed sequentially on the input table to produce the final
answer.

%\[
%a = \mathrm{Exec}(p, T).
%\]

\subsection{Instruction Design}
To enable models to generate executable and interpretable programs for table reasoning, we have developed a structured instruction format that explicitly aligns reasoning steps with code generation in Pandas. Rather than supervising reasoning and code generation as separate processes, our instructions require a unified output. Consequently, high-level planning, intermediate reasoning, and specific dataframe operations are integrated within a single Python function.

As illustrated at point (c) in Figure~\ref{fig:framework}, all outputs adhere to a standardized template: the function accepts a Pandas DataFrame as input and returns the final answer encapsulated in a list. Standardizing the output structure, rather than individual operations, minimizes variability unrelated to execution, aligns generation with code-centric pretraining data, and facilitates reliable execution-based evaluation.

Point (a) in Figure~\ref{fig:framework} demonstrates the commented reasoning structure employed in this approach, in which stepwise reasoning and question-specific constraints are articulated as structured comments closely integrated with executable Pandas operations. Each solution is required to begin with a mandatory one-line planning comment (e.g., \texttt{\# PLAN}, \texttt{\# FILTER}, \texttt{\# PARSING}, \texttt{\# AGGREGATE}) that succinctly summarizes the overall strategy for addressing the table question. The ellipsis at point (c) following \texttt{\# PLAN:} serves as a placeholder, indicating that the model must explicitly generate a planning statement rather than omitting or generalizing this step.

Subsequent reasoning steps are included only when necessary and are presented as concise, operation-level comments (e.g., \texttt{\# FILTER}, \texttt{\# PARSING}, \texttt{\# AGGREGATE}) that directly precede the corresponding DataFrame transformations. In addition to these reasoning steps, question-specific constraints, such as output quantity, ordering, and formatting, are encoded as explicit instruction-following requirements within the same commented structure. Treating these constraints as strict conditions within the generated program encourages the model to satisfy them directly in code, rather than relying on post hoc formatting. This approach reduces common failure cases in which computations are correct, but outputs do not meet the question specification.

As illustrated at point (b) in Figure 1, the instructions incorporate conventions for robust table execution that address common sources of noise, including mixed data types, numeric values stored as strings, and inconsistent date formats. The model is guided to apply safe parsing (e.g., \texttt{pd.to\_numeric(..., errors='coerce')}), handle missing values, and exclude summary rows unless explicitly required. These constraints encourage programs to operate on properly typed DataFrame representations rather than raw strings, thereby enhancing robustness to real-world table noise.

\begin{comment}
\begin{algorithm}[!t]
\caption{Constructing Training Data}
\label{alg:data_construction}
\small
\begin{algorithmic}[1]
\REQUIRE WikiTQ training and validation set
\ENSURE Commented Pandas-based programs

\FORALL{$(T, q, a)$ in WikiTQ}
    \STATE Generate commented program $p$ with mandatory \texttt{\# PLAN}
    \STATE $a_p \leftarrow \textsc{Execute}(p, T)$
    \IF{$a_p \neq a$}
        \STATE Refine $p$ using $(T, q, p, a_p, a)$
    \ENDIF
    \IF{$p$ trivially copies $a$}
        \STATE \textbf{discard} $p$
    \ELSE
        \STATE Add $p$ to the training set
    \ENDIF
\ENDFOR
\end{algorithmic}
\end{algorithm}

\begin{algorithm}[!t]
\caption{End-to-End Answer Selection}
\label{alg:e2e_selector}
\begin{algorithmic}[1]
\REQUIRE Table $T$, question $q$
\ENSURE Final predicted answer $\hat{a}$

\STATE $p \sim \pi_\theta(T, q)$
\COMMENT{Generate a commented step-by-step program}

\STATE $a_p \leftarrow \textsc{Execute}(p, T)$
\COMMENT{Execute the program once to obtain the code-based answer}

\STATE $(r_{\text{e2e}}, a_{\text{e2e}}) \leftarrow \textsc{Run TableR1}(T, q)$
\COMMENT{Run the TableR1 model to obtain reasoning trace and answer}

\STATE $\hat{a} \leftarrow \textsc{SelectAnswer}\big(
      (p, a_p),
      (r_{\text{e2e}}, a_{\text{R1}}),
      q \big)$
\COMMENT{Select the final answer based on complementary reasoning artifacts}

\RETURN $\hat{a}$
\end{algorithmic}
\end{algorithm}
\end{comment}

\subsection{Constructing Training Data}
Training data are constructed from the WikiTableQuestions (WikiTQ)~\citep{pasupat2015wtq} training and validation splits, which provide table–question–answer triples without intermediate reasoning or executable programs. To obtain commented Pandas-based solutions, Python functions are generated using Qwen2.5-Coder-32B-Instruct, with a mandatory high-level planning step (\texttt{\# PLAN}) and, when necessary, optional step-wise reasoning comments. The prompting protocol explicitly enforces question constraints and robustly manages heterogeneous or noisy table entries.

Figure~\ref{fig:dataset_generation} illustrates the overall dataset generation pipeline. For each table–question pair, the model first produces a commented program in a single pass. As this initial generation may result in incorrect or incomplete solutions, the pipeline incorporates a lightweight error-guided refinement procedure. When a failure is detected, the model is provided with the original table and question, the incorrectly generated code, its execution output, and the corresponding golden answer, and is prompted to generate a revised solution in the same commented format. To prevent trivial memorization, solutions that merely replicate the golden answer are excluded, and the resulting refined programs are used for supervised fine-tuning.

\subsection{Answer Selector}
\label{sec:selector}
Figure~\ref{fig:Answer selector} illustrates the overall structure of the proposed answer selection mechanism, which integrates code-based execution with the end-to-end TableQA model Table-R1~\citep{wang2025tabler1}. As shown in the figure, two candidate answers are generated for each table–question pair: one obtained by executing the Pandas program produced by the proposed method, and the other generated by Table-R1 using natural language reasoning. Motivated by the complementary strengths of these two approaches, a lightweight selector is trained to choose the final answer based solely on the input table and question. To construct training data for the selector, both methods are applied to the WikiTQ training split, and only examples in which at least one of the two predictions matches the golden answer are retained, while cases where both methods fail are discarded.

\begin{comment}
Algorithm~\ref{alg:e2e_selector} provides an overview of the end-to-end inference procedure employed in this study, which combines the proposed commented-code-based execution with the recent reasoning-centric model Table-R1~\citep{wang2025tabler1}. Given two candidate answers $a_p$ and $a_{\text{R1}}$, the final prediction is defined as
\[
\hat{a} = \arg\max_{a \in \{a_p, a_{\text{R1}}\}} s_\phi(a \mid T, q),
\]
where $s_\phi$ represents a lightweight answer selection model. This model is trained to score each candidate from commented-code-based execution and the reasoning-centric model Table-R1 based on the input table and question.
\end{comment}

\section{Experiments}

\subsection{Datasets}
The framework is evaluated on the WikiTableQuestions (WikiTQ) benchmark~\citep{pasupat2015wtq}, which is well aligned with the proposed code-based reasoning approach. Datasets featuring free-form answers, such as FeTaQA~\citep{nan2021fetaqa}, are not suitable because mapping arbitrary text to executable code presents challenges that extend beyond table reasoning.
%Existing benchmarks with larger structured tables are lacking, which would better test the scalability of program-based reasoning.

\subsection{Experimental Setup}
The proposed approach is evaluated on the WikiTQ test split. Tables are serialized for model input, and execution is conducted on a separately instantiated DataFrame that mirrors the original table. To ensure stable execution, lightweight, rule-based normalization is applied, such as removing thousands separators, normalizing special characters, and standardizing column names. This process preserves semantics while improving numeric parsing. For code generation, a diverse set of instruction-tuned and code-specialized models is evaluated, including Llama-3.1-8B-Instruct~\citep{dubey2024llama3}, Qwen2.5-7B-Instruct, Qwen2.5-Coder-7B/14B/32B-Instruct~\citep{qwen2024qwen25}, DeepSeek-Coder-7B-Instruct~\citep{guo2024deepseekcoder}, and Mistral-7B-Instruct~\citep{jiang2023mistral}, to assess the robustness of the our framework across model families and scales. For answer selection, lightweight models from the Qwen3 series~\citep{qwen2025qwen3} are primarily adopted, reflecting a practical scenario in which a small selector is paired with a more powerful code generator. Llama-3.1-8B-Instruct is also included to examine cross-family generalization.

\subsection{Inference and Evaluation Metrics}
During inference, each model generates executable Python code, which is then executed to obtain a final answer. Both Exact Match (EM) and Fuzzy Match (FM) metrics are reported, with FM serving as the primary evaluation metric unless otherwise specified. They are defined as follows:
\begin{align}
\mathrm{EM}(q, T) &= \mathbb{I}\big( f(T, q) = a \big) \label{eq:em} \\[6pt]
\mathrm{FM}(q, T) &= \mathbb{I}\big( \mathrm{match}(f(T, q), a) = 1 \big) \label{eq:fm}
\end{align}
where $q$ denotes the question, $T$ the table, $a$ the golden answer, $f(q,T)$ the model prediction obtained via code execution, and $\mathbb{I}(\cdot)$ the indicator function.

The WikiTQ dataset is crowdsourced from Wikipedia tables, resulting in golden answers with non-uniform surface forms. This variability includes differences in name qualifiers and benign formatting mismatches introduced by execution-based parsing, such as date normalization. Consequently, EM may incorrectly penalize semantically correct predictions. For completeness, EM results for all models are provided in Appendix~\ref{appendix_metrics}. FM accounts for minor surface-form variations while preserving semantic equivalence, providing a more faithful evaluation of reasoning correctness. We therefore adopt FM as the primary metric.

\begin{table}[!t]
\centering
\small
\resizebox{\columnwidth}{!}{
\begin{tabular}{lc}
\toprule
\textbf{Model} & \textbf{Accuracy (\%)} \\
\midrule
Llama-3.1-8B-Instruct              & 69.3 \\
Qwen2.5-7B-Instruct                & 69.3 \\
Qwen2.5-Coder-7B-Instruct          & 70.9 \\
DeepSeek-Coder-7B-Instruct-v1.5    & 68.0 \\
Mistral-7B-Instruct                & 66.9 \\
Qwen2.5-Coder-14B-Instruct         & 72.1 \\
Qwen2.5-Coder-32B-Instruct (qLoRA) & 71.2 \\
\bottomrule
\end{tabular}
}
\caption{
Performance of the proposed framework on the WikiTQ test set, evaluated using the FM accuracy.
}
\label{tab:ours_main_results}
\end{table}

\subsection{Results of the Proposed method}
Table~\ref{tab:ours_main_results} presents the performance of the proposed framework across instruction-tuned models on the original WikiTQ test set, without correcting known annotation errors. Including proposed method results in substantial performance improvements, even at the 7B parameter scale. For example, Qwen2.5-Coder-7B-Instruct achieves 70.9\% Fuzzy Match accuracy, indicating that embedding reasoning comments into Pandas-based code generation enhances table reasoning, yielding compact, code-oriented models, and surpassing prior program-based approaches at comparable scales. These improvements are maintained as model capacity increases. Qwen2.5-Coder-14B-Instruct achieves the highest overall performance, suggesting that the proposed framework scales effectively with model size.

\begin{table*}[!t]
\centering
\small
\begin{tabular}{l l c}
\hline
\textbf{Method} & \textbf{Base Model} & \textbf{Accuracy (\%)} \\
\hline
Table-R1 (Wang et al., 2025) & Llama-3.1-8B-Instruct & 81.7 \\
TabLaP (Wang et al., 2024a) & GPT-3.5-Turbo & 76.6 \\
SynTQA (GPT) (Zhang et al., 2024) & GPT-3.5-Turbo & 74.4 \\
Mix-SC (Liu et al., 2023) & GPT-3.5-Turbo & 73.6 \\
SynTQA (RF) (Zhang et al., 2024) & GPT-3.5-Turbo & 71.6 \\
CABINET (Patnaik et al., 2024) & Hybrid pipeline (no fixed LLM) & 69.1 \\
Repanda (Chegini et al., 2025) & DeepSeek-Coder-7B-Instruct-v1.5 & 67.6 \\
Chain-of-Table (Wang et al., 2024b) & GPT-3.5-Turbo & 67.3 \\
Tab-PoT (Xiao et al., 2024) & GPT-3.5-Turbo & 66.8 \\
\hline
Proposed method & Qwen2.5-Coder-14B-Instruct & 72.1 \\
\textbf{Proposed method} & \textbf{Qwen2.5-Coder-7B-Instruct} & \textbf{70.9} \\
\quad + \textbf{Answer Selector with Table-R1} & \quad + \textbf{Qwen3-4B-Instruct-2507} & \textbf{84.3} \\
\hline
\end{tabular}
\caption{Comparison with prior TableQA methods on the WikiTQ test set. All results are reported using Fuzzy Match accuracy, following standard evaluation practice in the literature.}
\label{tab:comparison_wikitq}
\end{table*}

\subsection{Comparison with Prior Methods}
Table~\ref{tab:comparison_wikitq} compares the proposed approach with representative prior TableQA methods on the WikiTQ benchmark, many of which rely on large, proprietary models such as GPT-3.5-Turbo. Despite relying on substantially smaller open-source models, the proposed comment reasoning method demonstrates competitive performance. Notably, Qwen2.5-Coder-7B-Instruct surpasses Repanda, the strongest prior program-based approach at a comparable scale. This improvement is attributed to the explicit decomposition of DataFrame operations into ordered, multi-line code with embedded reasoning, which stabilizes filtering, parsing, and aggregation decisions during execution.

Reasoning-centric models such as Table-R1 achieve strong results by directly reasoning over tables, but rely on implicit numerical reasoning through token generation, which can be fragile for aggregation-heavy queries. The code-based approach presented here is therefore complementary, as it provides deterministic numerical execution and improved interpretability.

Overall, while proprietary-model-based systems continue to lead in absolute accuracy, the results indicate that structured, code-grounded reasoning substantially narrows the performance gap in the small-model regime, offering a practical alternative for real-world TableQA.

\begin{figure}[!t]
    \centering
    \includegraphics[width=0.7\columnwidth]{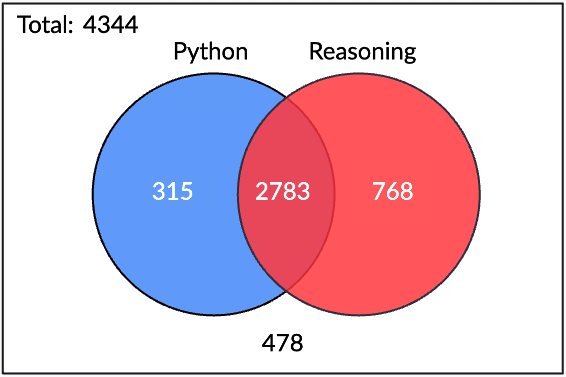}
    \caption{Performance breakdown of the proposed Python-based TableQA reasoning model and end-to-end TableQA reasoning model, Table-R1.}
    \label{fig:performance breakdown}
\end{figure}

\begin{figure}[!t]
    \centering
    \includegraphics[width=\columnwidth]{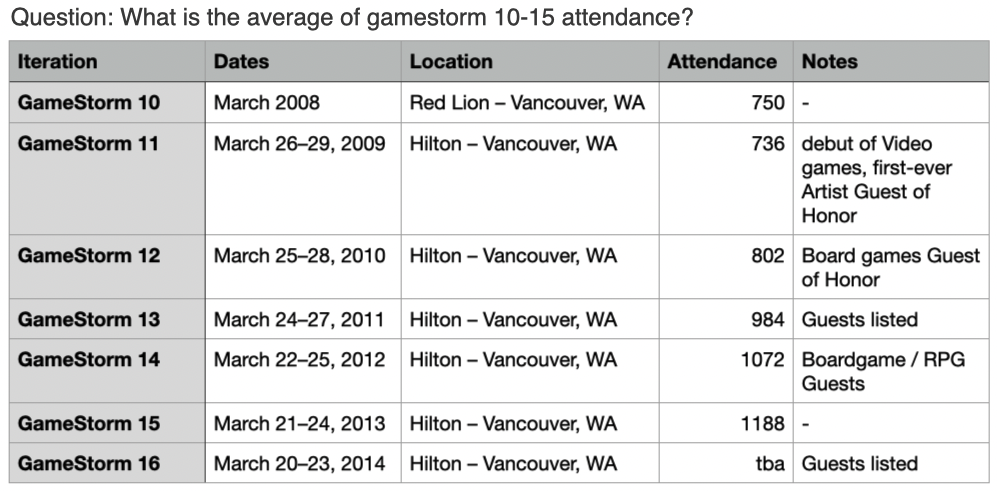}
    \caption{Testset 2521}
    \label{fig:test2521}
\end{figure}

\subsection{Code-Based Commented Reasoning vs. Table-R1}
Figure~\ref{fig:performance breakdown} presents a comparison of prediction outcomes on the WikiTQ test set between the proposed Python-based approach and Table-R1. Both models correctly answer 2,783 questions; however, 315 questions are solved exclusively by the proposed approach, and 768 are addressed only by Table-R1.

Focusing on cases where the proposed Python-based approach succeeds, but Table-R1 fails, we observe recurring error patterns, most notably numerical operation failures. As shown in Figure~\ref{fig:test2521}, the end-to-end model fails to correctly aggregate numeric values despite all required information being present, whereas the proposed approach performs deterministic DataFrame operations and produces the exactly correct answer (Table~\ref{tab:average_failure_summary}).

\begin{table}[!t]
\centering
\small
\begin{tabular}{lr}
\toprule
\textbf{Method} & \textbf{Output} \\
\cmidrule(lr){1-2}
Table-R1 (End-to-end TableQA) & 918.67 \\
Proposed method (Code-based Reasoning)           & 922 \\
\cmidrule(lr){1-2}
Golden Answer                    & 922 \\
\bottomrule
\end{tabular}
\caption{
Comparison of model outputs on a numerical averaging question. Table-R1 fails to correctly divide value,
while the proposed execution-based approach produces the exactly correct answer.
}
\label{tab:average_failure_summary}
\end{table}

In addition to numerical errors, Table-R1 exhibits failures resulting from empty cells and reasoning instability, where correct intermediate interpretations are not maintained. Additional examples are presented in Appendix~\ref{failure_analysis}. These patterns reveal structural limitations inherent in reasoning-centric approaches and underscore the need for complementary integration with program-based execution.

\begin{table}[!t]
\centering
\small
\begin{tabular}{l c}
\hline
\textbf{Answer Selector Model} & \textbf{Accuracy} \\
\hline
Llama-3.1-8B-Instruct & 83.9 \\
Qwen3-8B & 84.2 \\
Qwen3-4B-Instruct-2507 & 84.3 \\
Qwen3-1.7B & 84.1 \\
Qwen3-0.6B & 83.6 \\
\hline
\end{tabular}
\caption{Final system accuracy after answer selection, combining the proposed code-based reasoning model and Table-R1 using selector models of varying sizes.}
\label{tab:answer_selector}
\end{table}

\subsection{Effect of the Answer Selector}
Due to complementary failure patterns observed in the proposed code-based execution and Table-R1, the two approaches are integrated via a lightweight answer selection module as described in Section~\ref{sec:selector}. Table~\ref{tab:answer_selector} presents the final accuracy achieved by selecting between executable code outputs and end-to-end predictions.

Answer selection consistently outperforms each individual approach. Integrating the commented code-based model with Table-R1 results in substantial improvements, even when using a compact selector model. For example, Qwen3-4B-Instruct-2507 achieves 84.3\% accuracy, and smaller selectors such as Qwen3-1.7B demonstrate comparable performance.

The selector is not an oracle. In 2.03\% of cases, correct code-based answers are overridden by incorrect Table-R1 predictions, and the reverse occurs in 1.20\% of cases. These errors indicate that answer selection remains a bottleneck and suggest that confidence- or uncertainty-aware selection could further improve robustness.

Collectively, these results indicate that the answer selector serves as a straightforward yet effective mechanism for integrating code-based TableQA reasoning into an end-to-end TableQA model, yielding a more robust TableQA reasoning system than either method alone.

\begin{table}[!t]
\centering
\resizebox{0.8\linewidth}{!}{
\begin{tabular}{lcc}
\toprule
\textbf{Model} & \textbf{Original} & \textbf{Corrected} \\
\midrule
Llama-3.1-8B-Instruct & 69.3 & 70.2 \\
Qwen2.5-Coder-7B-Instruct & 70.9 & 71.9 \\
Qwen2.5-7B-Instruct & 69.3 & 70.2 \\
DeepSeek-Coder-7B-Instruct & 68.0 & 69.2 \\
Mistral-7B-Instruct & 66.9 & 67.8 \\
Qwen2.5-Coder-14B-Instruct & 72.1 & 73.0 \\
Repanda & 67.6 & - \\
Table-R1 & 81.7 & 82.7 \\
\midrule
Proposed method + Table-R1 & 84.3 & 85.3 \\
\bottomrule
\end{tabular}
}
\caption{A performance comparison was conducted using original and corrected annotations on the WikiTQ test set, measured by Fuzzy Match accuracy. The answer selector result represents a system-level combination of the proposed code-based TableQA model and the end-to-end TableQA model, Table-R1. Results for Repanda are reported from the original publication, since the model and training data are not publicly accessible.}
\label{tab:annotation_correction}
\end{table}

\subsection{Annotation Errors in the WikiTQ Test Set}

Instances were observed in which both the proposed code-based TableQA model and the end-to-end TableQA model Table-R1 generated identical predictions that did not align with the golden answers, indicating potential annotation errors in the WikiTQ test set rather than model shortcomings.

To quantify the impact of this noise, a subset of test instances with clear and unambiguous annotation errors was manually identified, and model performance was re-evaluated using corrected answers. As shown in Table~\ref{tab:annotation_correction}, all models demonstrate consistent performance improvements following correction, indicating that dataset noise represents a significant source of residual errors. For fairness and comparability with previous studies, all main results are reported using the original WikiTQ annotations. Specific examples of annotation errors are presented in Appendix~\ref{annotation_errors}.

\section{Conclusion}

This paper introduces a commented reasoning framework for TableQA that bridges the gap between end-to-end generation and executable programming. By decomposing complex table reasoning into explicit, multi-line Pandas operations paired with structured natural language comments (such as \# PLAN, \# FILTER, and \# AGGREGATE), the proposed system achieves both high reliability and human interpretability. In contrast to previous program-aided methods that rely on opaque, single-line queries, this approach grounds each step of the reasoning process in a deterministic symbolic execution engine, thereby ensuring numerical exactness and robustness to noise in real-world tabular data.

Experimental results on the WikiTQ benchmark demonstrate that the proposed framework consistently improves performance across various instruction-tuned and code-specialized models. Notably, the method using Qwen2.5-Coder-7B-Instruct achieves an accuracy of 70.9\%, significantly outperforming the state-of-the-art code-based model, Repanda. Furthermore, code-based TableQA reasoning and end-to-end TableQA reasoning exhibit complementary error patterns. Integrating the proposed code-based method with the state-of-the-art end-to-end TableQA reasoning model, Table-R1, through a lightweight answer selector yields a state-of-the-art accuracy of 84.3\%.

Grounding table reasoning in step-wise, commented code provides a scalable foundation for addressing intensive numerical tasks and unstructured table formats. This approach narrows the performance gap for smaller, open-source models and offers a transparent, verifiable mechanism for TableQA in practical, real-world applications.

%We proposed a commented reasoning framework for table question answering that improves the reliability and interpretability of Python-based program generation. By decomposing reasoning into explicit, step-wise dataframe operations with aligned comments, our approach avoids forcing the entire reasoning process into a single pandas expression, while remaining efficient and consistent with practical coding practices. Experiments on WikiTableQuestions demonstrate consistent improvements across multiple instruction-tuned models.

%We further show that code-based method and end-to-end method exhibit complementary error patterns. Combining our approach with a strong end-to-end TableQA model via a lightweight answer selection mechanism yields additional gains, reaching up to 84.3\% accuracy on WikiTQ. Overall, step-wise, code-grounded reasoning provides a scalable and interpretable foundation for robust table reasoning in noisy and numerically intensive settings.

\section{Limitations}

One limitation of our work is that the proposed approach is evaluated primarily on WikiTQ, a benchmark well suited for executable, python-based table reasoning. This choice reflects our focus on TableQA settings where explicit program execution can provide clear advantages, such as non-trivial numerical operations and multi-step data transformations. In contrast, for very small tables or questions requiring minimal computation, end-to-end reasoning models are often
sufficient and may even be preferable. As a result, our evaluation does not fully characterize
the behavior of the proposed method across all TableQA regimes. Exploring how code-based reasoning interacts with table size and task complexity remains an interesting direction for future work.

\section*{Impact Statement}

This work contributes to the field of machine learning by improving the reliability and interpretability of table question answering systems through explicit, executable reasoning. More accurate and transparent table reasoning models may benefit real-world applications that rely on numerical decision-making over structured data, such as data analysis, report generation, and information access.

At the same time, like other machine learning systems, models developed using this approach may inherit biases or errors present in underlying data or annotations. However, by making intermediate reasoning steps explicit and executable, our framework may facilitate error analysis and model auditing, which can help mitigate such risks. Overall, we do not foresee significant adverse societal impacts specific to this work.

\nocite{langley00}

\bibliography{references}
\bibliographystyle{icml2026}

%%%%%%%%%%%%%%%%%%%%%%%%%%%%%%%%%%%%%%%%%%%%%%%%%%%%%%%%%%%%%%%%%%%%%%%%%%%%%%%
%%%%%%%%%%%%%%%%%%%%%%%%%%%%%%%%%%%%%%%%%%%%%%%%%%%%%%%%%%%%%%%%%%%%%%%%%%%%%%%
% APPENDIX
%%%%%%%%%%%%%%%%%%%%%%%%%%%%%%%%%%%%%%%%%%%%%%%%%%%%%%%%%%%%%%%%%%%%%%%%%%%%%%%
%%%%%%%%%%%%%%%%%%%%%%%%%%%%%%%%%%%%%%%%%%%%%%%%%%%%%%%%%%%%%%%%%%%%%%%%%%%%%%%
\newpage
\appendix
\onecolumn

\section{Failure Analysis of End-to-End Reasoning Models}
\label{failure_analysis}
% =========================
% A.1 Numerical Operation Failure
% =========================

\FloatBarrier
\subsection{Numerical operation failure}
\begin{figure}[H]
    \centering
    \includegraphics[width=0.8\textwidth,keepaspectratio]{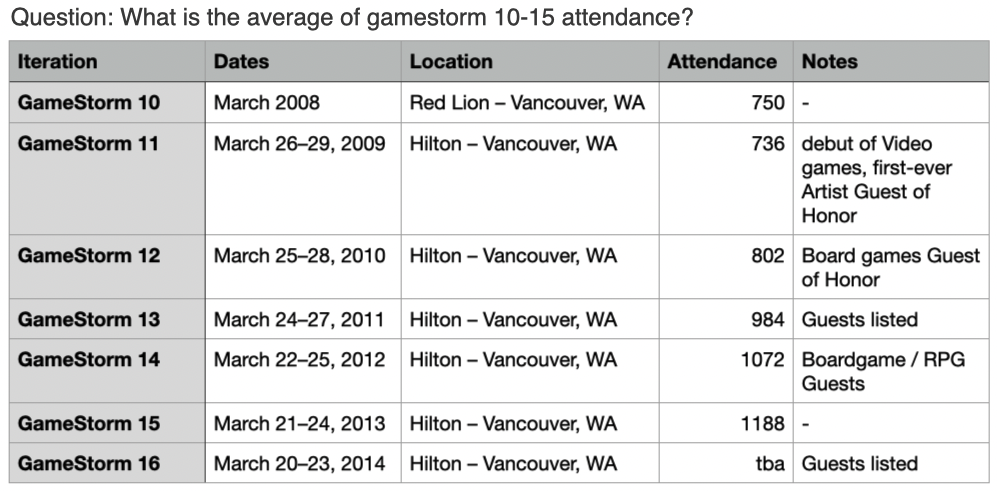}
    \caption{Testset 2521}
    \label{fig:appendix_testset_2521}
\end{figure}

In the example shown in Figure~\ref{fig:appendix_testset_2521} and Table~\ref{tab:appendix_numerical_failure}, the end-to-end reasoning model (Table-R1) correctly identifies the relevant rows and successfully computes the sum of the attendance values for GameStorm 10 through 15. However, the final averaging step is carried out purely through implicit token-level reasoning rather than an explicit, verifiable arithmetic operation. As a result, the division step lacks a deterministic computational mechanism, leading to a rounding error in the final output (918.67 instead of the correct value 922).

This failure highlights a structural limitation of end-to-end reasoning approaches: while intermediate quantities such as partial sums may be correctly inferred, arithmetic operations such as division are not grounded in an executable procedure and therefore cannot guarantee numerical exactness. In contrast, our code-based approach explicitly performs both aggregation and division through pandas dataframe operations, ensuring that the mean is computed deterministically according to the underlying data semantics. This example illustrates how execution-based reasoning mitigates numerical instability by delegating arithmetic to an explicit computational process, rather than relying on approximate symbolic reasoning within the language model.

\begin{table}[t]
\centering
\small
\begin{tabular}{p{1.7cm} p{12.0cm} p{1.0cm}}
\hline
\textbf{Method} & \textbf{Reasoning / Code Snippet} & \textbf{Output} \\
\hline

Table-R1 &
\begin{minipage}[t]{0.98\linewidth}
\ttfamily\small
\textless think\textgreater

Okay, let’s see. The question is asking for the average attendance of GameStorm 10 through 15. So first, I need to look at the table provided and extract the attendance numbers for each of those years.
Looking at the table rows, GameStorm 10 has an attendance of 750, GameStorm 11 has 736, GameStorm 12 has 802, GameStorm 13 has 984, GameStorm 14 has 1072, and GameStorm 15 has 1188. Since the question is about GameStorm 10 to 15 inclusive, that corresponds to six events in total.

Now I need to sum these attendance values and divide by the number of events. Let me add them step by step: 750 plus 736 equals 1486. Adding 802 gives 2288. Adding 984 gives 3272. Adding 1072 gives 4344. Adding 1188 gives a total sum of 5532.

The total sum is 5532, and since there are six events, I divide 5532 by 6 to
compute the average. Doing the division, 5532 divided by 6 equals 918.666\ldots

[...]

After reconsidering the calculation and the expected output format, the average
attendance for GameStorm 10 through 15 is therefore 918.67.

\textless /think\textgreater

\textless answer\textgreater
\{
  "answer": ["918.67"]
\}
\textless /answer\textgreater
\end{minipage}
&
918.67 \\
\hline

Ours &
\begin{minipage}[t]{0.95\linewidth}
\ttfamily\small
def solution(df):\newline
% \newline
\hspace*{1em}\# PLAN: Filter the rows for GameStorm 10--15 and compute the average attendance.\newline
\hspace*{1em}\# FILTER: Select relevant iterations.\newline
\hspace*{1em}filtered\_df = df[(df['Iteration'] >= 'GameStorm 10') \& (df['Iteration'] <= 'GameStorm 15')]\newline
% \newline
\hspace*{1em}\# PARSING: Convert Attendance to numeric.\newline
\hspace*{1em}filtered\_df['Attendance'] = pd.to\_numeric(filtered\_df['Attendance'], errors='coerce')\newline
% \newline
\hspace*{1em}\# AGGREGATE: Compute the mean.\newline
\hspace*{1em}average\_attendance = filtered\_df['Attendance'].mean()\newline
% \newline
\hspace*{1em}\# ANSWER\newline
\hspace*{1em}return [average\_attendance]
\newline
\end{minipage}

&
922 \\
\hline

Gold Answer & -- & 922 \\
\hline
\end{tabular}
\caption{
Numerical operation failure in end-to-end reasoning.
Despite producing a detailed reasoning trace, Table-R1 arrives at an incorrect
numerical result, while execution-based reasoning strictly follows dataframe
semantics.
}
\label{tab:appendix_numerical_failure}
\end{table}

% =========================
% A.2 Empty-Cell–Induced Column Misalignment
% =========================
\clearpage
\subsection{Empty-Cell--Induced Column Misalignment}

\begin{figure}[H]
    \centering
    \includegraphics[width=0.8\textwidth,,keepaspectratio]{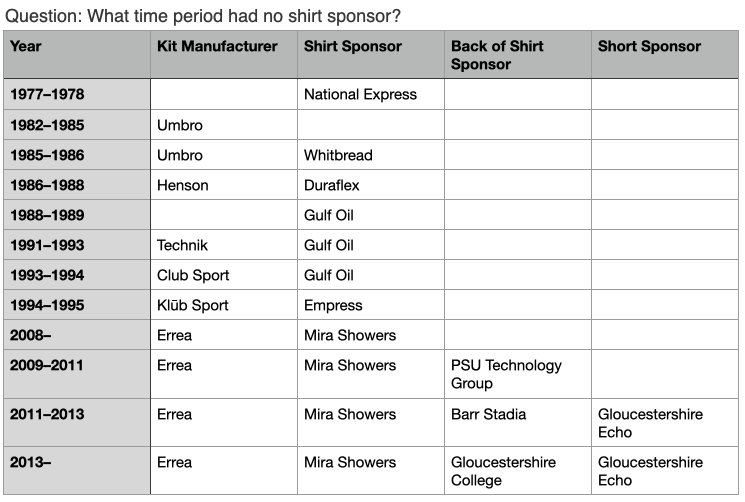}
    \caption{Testset 0009}
    \label{fig:appendix_testset_0009}
\end{figure}

The failure case in Figure~\ref{fig:appendix_testset_0009} and Table~\ref{tab:appendix_empty_cell_failure} illustrates a common issue arising from empty cells in serialized tables. When the table is converted into a text sequence for model input, columns are separated by the delimiter '\texttt{|}'. Empty cells are therefore represented as consecutive delimiters (i.e., '\texttt{||}'), which provides no explicit signal indicating the absence of a value. As a result, the end-to-end reasoning model fails to consistently recognize empty entries as valid missing values and instead implicitly associates them with neighboring columns or headers.

This ambiguity leads Table-R1 to incorrectly infer multiple time periods without a shirt sponsor, including rows where the empty cell does not correspond to the target column. Notably, during its reasoning trace, the model momentarily identifies the correct answer (``1982--1985'') after re-evaluating the table structure. However, this correct intermediate interpretation is not preserved throughout the remainder of the reasoning process and is ultimately overridden by an earlier, incorrect hypothesis. This behavior reflects a form of reasoning instability, where intermediate conclusions are not reliably maintained across longer reasoning chains.

In contrast, the execution-based approach operates directly on a structured dataframe, where empty cells are explicitly represented and column boundaries are unambiguous. By filtering rows based on the exact column semantics of the ``Shirt Sponsor'' field, the code-based method deterministically retrieves the correct time period. This example demonstrates how empty-cell ambiguity in text-based table representations can induce column misalignment errors in end-to-end models, while programmatic execution avoids such failures by preserving the original table structure.

\clearpage
\begin{table}[H]
\centering
\small
\begin{tabular}{p{2.5cm} p{9.5cm} p{2.5cm}}
\hline
\textbf{Method} & \textbf{Reasoning / Code Snippet} & \textbf{Output} \\
\hline

Table-R1 &
\begin{minipage}[t]{0.95\linewidth}
\ttfamily\small
\textless think\textgreater

Okay, let's see. The question is asking which time period had no shirt sponsor. So I need to look at the table provided and check the \"Shirt Sponsor\" column for any entries that are empty or missing. Looking at the table rows: Starting from the top, the first entry is 1977-1978. The Shirt Sponsor column is empty here. Then the next few entries have sponsors listed, like National Express, Umbro, Whitbread, etc. Then in 1988-1989, the Shirt Sponsor column is empty again. Then from 1991 onwards, there are entries again. Wait, let me check each row carefully. So the first row (1977-1978) has no Shirt Sponsor. \textbf{Then 1982-1985, Shirt Sponsor is empty.} Then 1985-1986 has Whitbread. Then 1986-1988 has Duraflex. Then 1988-1989, Shirt Sponsor is empty again. Then 1991--1993 has Gulf Oil. The rest of the rows from 1993 onwards have Shirt Sponsors listed. So the time periods with no shirt sponsor are 1977-1978 and 1988-1989. Let me double-check. For 1977-1978, the Shirt Sponsor column is blank. For 1988-1989, same thing. All other years have some sponsor listed. So those two periods are the answer.

\textless /think\textgreater

\{
  "answer": ["1977--1978", "1988--1989"]
\}
\end{minipage}
&
1977--1978, 1988--1989 \\
\hline

Ours (Commented Code) &
\begin{minipage}[t]{0.95\linewidth}
\ttfamily\small
def solution(df):\newline
\hspace*{1em}\# PLAN: Find rows with empty Shirt Sponsor entries.\newline
\hspace*{1em}\# FILTER\newline
\hspace*{1em}no\_sponsor = df[df['Shirt Sponsor'] == ''][['Year']]\newline
\hspace*{1em}\# ANSWER\newline
\hspace*{1em}return no\_sponsor['Year'].tolist()
\end{minipage}
&
1982--1985 \\
\hline

Gold Answer & -- & 1982--1985 \\
\hline
\end{tabular}
\caption{
Empty-cell--induced column misalignment.
End-to-end reasoning incorrectly associates empty cells with neighboring headers,
while execution-based reasoning preserves correct column semantics.
}
\label{tab:appendix_empty_cell_failure}
\end{table}

\clearpage
\section{Examples of Annotation Errors in WikiTQ}
\label{annotation_errors}
\subsection{Testset 0130}
\vspace*{-15pt}
\begin{figure}[H]
    \centering
    \includegraphics[width=0.6\textwidth,height=0.5\textheight,keepaspectratio]{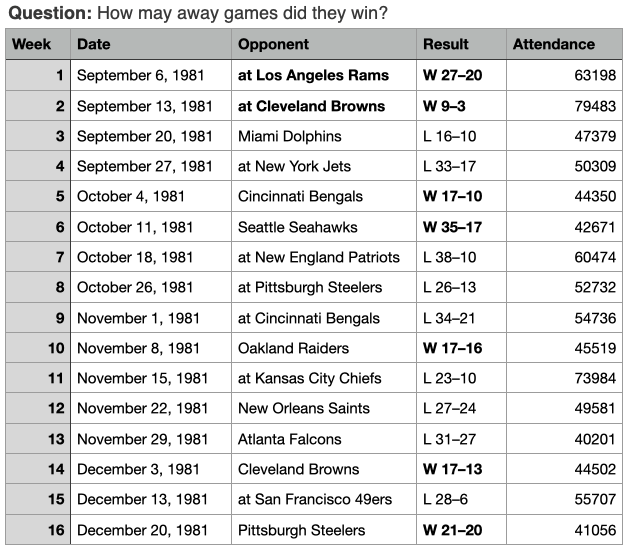}
    \caption{Testset 0130}
    \label{fig:appendix_testset_0130}
\end{figure}

In this instance, the annotated gold answer is incorrect. While the dataset annotation reports an answer of \textbf{7}, the correct answer is \textbf{2}. This discrepancy stems from an incorrect aggregation criterion used during annotation.
The question explicitly asks for the number of \emph{away wins}, but the annotated answer incorrectly counts all wins irrespective of match venue. In the table, away matches are unambiguously indicated by the prefix ``at'' in the \emph{Opponent} column, providing an explicit and machine-verifiable signal. When this indicator is properly taken into account, only two wins correspond to away games, yielding the correct answer of 2.
This example illustrates that the observed error is not due to model misreasoning, but rather to a misalignment between the question specification and the dataset annotation.

% \clearpage
\subsection{Testset 0184}
\begin{figure}[H]
    \centering
    \includegraphics[width=0.6\textwidth,keepaspectratio]{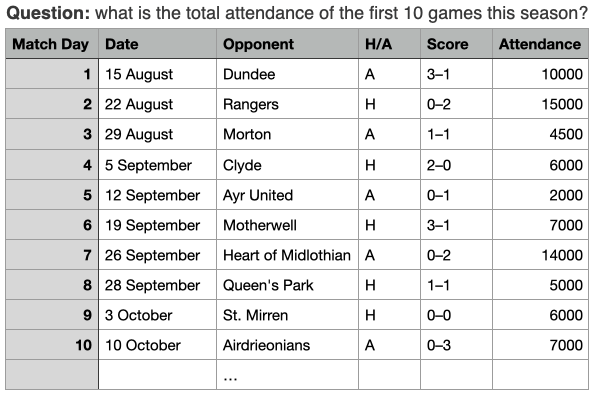}
    \caption{Testset 0184}
    \label{fig:appendix_testset_0184}
\end{figure}

The question asks for the total attendance over the first ten games of the season, and the aggregation criterion is clear and unambiguous. All required attendance values are explicitly provided in the table.
When summing the attendance for the first ten matches, the correct total is \textbf{76{,}500}, whereas the annotated gold answer reports \textbf{76{,}000}. Since there is no plausible alternative interpretation of either the question or the table entries, this discrepancy is attributable to a simple arithmetic error in the dataset annotation rather than to ambiguity or model misreasoning.
This example highlights that a portion of remaining errors in WikiTQ stems from annotation noise even in cases involving straightforward numerical aggregation.

% =========================
% Appendix X: Metrics
% =========================
\section{Exact Match and Fuzzy Match Results}
\label{appendix_metrics}
\begin{table}[H]
\centering
\small
\begin{tabular}{lcc}
\toprule
\textbf{Model} & \textbf{Exact Match (\%)} & \textbf{Fuzzy Match (\%)} \\
\midrule
Llama-3.1-8B-Instruct              & 65.2 & 69.3 \\
Qwen2.5-7B-Instruct                & 64.7 & 69.3 \\
Qwen2.5-Coder-7B-Instruct          & 66.0 & 70.9 \\
DeepSeek-Coder-7B-Instruct-v1.5    & 63.7 & 68.0 \\
Mistral-7B-Instruct                & 63.0 & 66.9 \\
Qwen2.5-Coder-14B-Instruct         & 67.4 & 72.1 \\
Qwen2.5-Coder-32B-Instruct (qLoRA) & 66.4 & 71.2 \\
\bottomrule
\end{tabular}
\caption{
Performance of the commented reasoning framework on the WikiTQ test set.
Fuzzy Match is used as the primary metric, with Exact Match reported for reference.
}
\end{table}

This appendix provides a detailed breakdown of Exact Match (EM) and Fuzzy Match (FM)
results for the WikiTQ test set.
As discussed in the main text, EM can be overly strict due to non-uniform surface
forms in the annotations.
We therefore report EM alongside FM to facilitate comprehensive evaluation and
comparison with existing TableQA benchmarks.

%%%%%%%%%%%%%%%%%%%%%%%%%%%%%%%%%%%%%%%%%%%%%%%%%%%%%%%%%%%%%%%%%%%%%%%%%%%%%%%
%%%%%%%%%%%%%%%%%%%%%%%%%%%%%%%%%%%%%%%%%%%%%%%%%%%%%%%%%%%%%%%%%%%%%%%%%%%%%%%

\end{document}